\documentclass[journal,onecolumn,10pt]{IEEEtran}
\usepackage{amsmath,amsfonts}
\usepackage{flushend}
\usepackage{multirow}
\usepackage{multicol}
\usepackage{algorithmic}
\usepackage{algorithm}
\usepackage{array}
\usepackage[caption=false,font=normalsize,labelfont=sf,textfont=sf]{subfig}
\usepackage{textcomp}
\usepackage{stfloats}
\usepackage{url}
\usepackage{verbatim}
\usepackage{graphicx}
\usepackage{color}
\usepackage{cite}
\begin{document}

\title{Semi-supervised segmentation of land cover images using nonlinear canonical correlation analysis with multiple features and t-SNE}

\author{Hong Wei, James  Xiao, Yichao Zhang, and Xia Hong 
\thanks{Hong Wei, Yichao Zhang, and Xia Hong are with the School of Mathematical, Physical and Computational Sciences,
 University of Reading, Reading,  RG6 6AY, UK. (h.wei/x.hong@reading.ac.uk, yichao.zhang@pgr.reading.ac.uk)}
 \thanks{James  Xiao is an independent researcher, Vancouver, BC, Canada, zifan.james.xiao@gmail.com.}}

\maketitle

\begin{abstract}
 Image segmentation is a clustering task whereby each pixel is assigned a cluster label. Remote sensing data usually consists of multiple bands of spectral images in which there exist semantically meaningful land cover subregions, co-registered with other source data such as LIDAR (LIght Detection And Ranging) data, where available. This suggests that, in order to account for spatial correlation between pixels, a feature vector associated with each pixel may be a vectorized tensor representing the multiple bands and a local patch as appropriate. Similarly, multiple types of texture features based on a pixel's local patch would also be beneficial for encoding locally statistical information and spatial variations, without necessarily labelling pixel-wise a large amount of ground truth, then training a supervised model, which is sometimes impractical.  In this work, by resorting to label only a small quantity of pixels, a new  semi-supervised segmentation approach is proposed. Initially, over all pixels, an image data matrix is created in high dimensional feature space. Then, t-SNE projects the high dimensional data onto 3D embedding. By using radial basis functions as input features, which use the labelled data samples as centres, to pair with the output class labels,  a modified canonical correlation analysis algorithm, referred to as RBF-CCA, is introduced which learns the associated projection matrix via the small labelled data set. The associated canonical variables, obtained for the full image, are applied by k-means clustering algorithm. The proposed semi-supervised RBF-CCA algorithm has been implemented on several remotely sensed multispectral images, demonstrating excellent segmentation results. 
 \end{abstract}


\section{Introduction}\label{S1}
 Image segmentation is an essential task in computer vision to partition an image into semantically meaningful regions with labelled (supervised) or without labelled (unsupervised) training samples. Obtaining dense image annotations pixel-wise is particularly difficult and expensive for remotely sensed images with complex scenes containing multiple categories, such as urban areas. It is desirable to utilise unsupervised image segmentation, which is a longstanding challenge in computer vision. Over the years, numerous methods have been developed to tackle the problem, from early approaches of clustering using statistical and mathematical models \cite{MacQueen1967,Caillol1993}, graph cut-related methods \cite{Shi2000, Zhu2019, Dutta2019}, evidence-based approaches \cite{Cao2012,Yang2020}, to more advanced deep learning techniques \cite{Kim2020, Zadaianchuk2023}. Researchers also attempt to adapt domain knowledge into deep learning based unsupervised image segmentation \cite{Wu2022,Liu2022,Wang2022} or use image labels in pixel-level segmentation \cite{Jing2019}.  However, these methods may be highly inconsistent with the true partition due to the lack of correspondence between samples and regions. To avoid using a large amount of labelled samples, semi-supervised approaches for image segmentation become popular as a research area \cite{Cai2023}, in which only a small set of labelled samples is needed to achieve a relatively high performance. 
 
\begin{figure*}[!h]
   \centering
\includegraphics[width=0.7\textwidth, height= 0.4\textwidth]{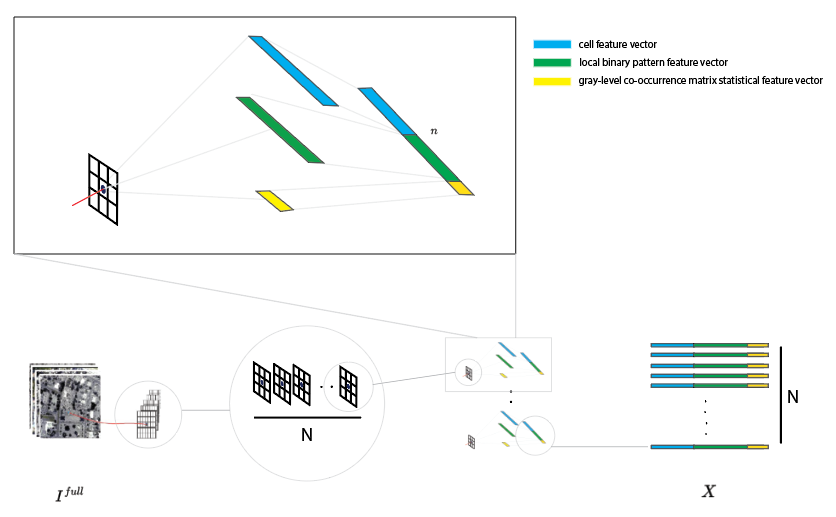 }
    \caption{Illustrative diagram of constructing data $\boldsymbol{X}$ for t-SNE then RBF-CCA from a remote sensing image $I^{full}$, where each pixel is mapped onto a feature vector consisting of multiple spatial features (for details, see  Section \ref{S4.2}).}
    \label{Figure:1}
\end{figure*}

In this study, the segmentation problem is formulated as a semi-supervised clustering approach, over the 3-D t-SNE embedding~\cite{Maaten2008} derived from a high-dimensional feature space mapped from pixels.  As illustrated in Figure \ref{Figure:1},   given a set of co-registered images,  any pixel in each band of original data is mapped onto a feature space, composed of a vectorized local patch, together with a set of image textural feature vectors \cite{Haralick1973,Ojala1994} also based on local patches centred at the pixel. The t-distributed stochastic neighbor embedding (t-SNE) algorithm, proposed by van der Maaten and Hinton, outlined in Section \ref{S3.1}, is a state-of-art technique for dimensionality reduction and visualization for a wide range of applications~\cite{WangYingfan2021}. Cai and Ma investigated the theoretical foundations of t-SNE, which can provide theoretical guidance for its application and help select its tuning parameters for various applications~\cite{CaiTony2022}. This is the foundation to build the semi-supervised clustering method, which aims to utilise a limited number of labelled data to predict a cluster's label in unlabelled data by training a clustering model using both unlabelled and labelled data. 

The segmenting process is based on a novel  algorithm, referred to as RBF-CCA (Radial Basis Function - Canonical Correlation Analysis), with data from t-SNE embedding of pixel based image textural feature vectors. CCA was originally proposed by Hotelling in the seminal works \cite{Hotelling1935,Hotelling1936}. It finds the best predictor among the linear functions from each set by maximizing the correlation coefficient between two sets, using spectral analysis. In an attempt to increase the flexibility of CCA for nonlinear relationships between two random variables,  kernel CCA  has been introduced~\cite{Akaho2001}. The proposed semi-supervised clustering algorithm can be viewed as a generalized spectral clustering algorithm, due to that CCA is based on eigen-decomposition of data matrices.

Our contributions to knowledge are listed below:
 \begin{enumerate}
 \item  A novel  pixel-based  multiple spatial feature vector has been proposed in order to fully exploit  spatially-resolved multi-band  image self-description for remotely sensed land cover imagery.
  \item  The proposed algorithm operates over 3D embedding via t-SNE aimed at clustering of high dimensional data sets generated by a novel pixel based feature vector.
    \item  In order to provide approximation of the nonlinear  shapes of  t-SNE  clusters,  radial basis functions (RBF) using labelled data input features, are employed as the first set of variable of RBF-CCA,  the class labels become  indicative variables as the second set of variables in CCA.
    \item The proposed algorithm transfers knowledge from a small set of labelled data, via projection of the associated canonical variables, to a large quantity of unlabelled data, to which k-means clustering is applied as used by a spectral clustering algorithm~\cite{UlrikeLuxburg2007}. 
    \item We discover that the resulting final   clusters are capable of matching extremely well to semantically meaningful image regions, such as buildings, trees, impervious surface, and low vegetation.  
\end{enumerate}
The rest of the paper is organized as follows. The related work with literature review is presented in Section \ref{S2}.  Section \ref{S3} introduces mathematical methods of  t-SNE embedding and canonical correlation analysis. Section \ref{S4} initially introduces the proposed pixel-based  multiple spatial feature generation from multi-band land-cover images, followed by the proposed semi-supervised algorithm based on t-SNE and RBF-CCA for  segmentation.  In Section \ref{S5}, experiments and results are presented and discussed. Finally, Section \ref{S6} concludes the work.

\section{Related work}\label{S2}
The concept of image segmentation by using clustering approaches   has been explored over time. Different learning approaches are developed, such as graph-based methods \cite{Shi2000, Ng2001, Cour2005, Zhang2008, Zhu2022, Zhong2023}, where spectral clustering algorithms are heavily exploited. Data clustering, including spectral clustering for image segmentation, is an unsupervised classification paradigm which divides observed data into different subsets (clusters). Yet, purely unsupervised spectral image segmentation often fails to achieve satisfactory performance due to that no truth labels are used for training.  In some cases, extra information is provided in the clustering output space to enhance the  clustering results. Clustering algorithms that combine complementary information  to supervise the cluster learning process are called semi-supervised clustering methods~\cite{Sugato2004,ZahraGhasemi2021}. For example, Kim, et al. \cite{Kim2013}, adopted a semi-supervised learning strategy in developing an efficient algorithm for image segmentation by employing spectral clustering in constructing a sparse multilayer graph with nodes of pixels and over-segmented regions from the image.

In \cite{Pont-Tuset2017}, Pont-Tuset, et al. developed a fast normalized cuts algorithm as the initial step to group pixels. They then proposed a multiscale combinatorial grouping algorithm to perform a hierarchical search for similar regions across different scales. In their research, scaled images were used to generate ultrametric contour maps in the segmentation process. Structured sparse subspace clustering was proposed by Li et al. \cite{Li2017}, in which each data point is expressed as a structured sparse linear combination of all other data points. Then, a joint optimisation framework was applied for learning both the affinity learned from the data and the segmentation implemented by spectral clustering. The concept of multi-view spectral clustering was applied in \cite{Luo2019} to deal with an NP-hard optimization problem which occurs in conventional spectral clustering approaches. The developed framework was evaluated for image segmentation along with other applications. In \cite{Jiao2020}, Jiao et al. applied spectral clustering to high-level features, which were outputs from an autoencoder network with its input a set of low-level image features over pre-generated non-overlapping superpixels. In their work, the low-level image features include colour, gradient, local binary pattern, and saliency features. With the unsupervised nature, the evaluation on two datasets showed IOU (Intersection over Union) about 0.6.     

The concept of few shots learning (FSL) is proposed~\cite{LiFeiFei2006, Wangyaqing2019} which refers to the learning  from a limited number of samples with supervised information.  This learning paradigm is desired since  large-scale supervised data can be expensive in many applications, e.g. in semantic segmentation to associate  a label or category with every pixel in an image~\cite{catalano2023}. For remote sensing image segmentation, Chen et al. attempted using semi-supervised approach to build a spectral-spatial classifier with affinity scoring to segment hyperspectral imagery \cite{Chen2015}, and Jiang et al. developed a semi-supervised segmentation method, in which superpixels are generated to construct the graph in a graph convolutional network for sea ice classification from remotely sensed synthetic aperture radar (SAR) imagery \cite{Jiang2022}.

Differently from the previous works, our study, categorised as a semi-supervised clustering approach, transforms high-dimensional image feature vectors into a 3D space by using t-SNE, and then a small set of labelled pixels is applied to RBF-CCA,  with unlabelled pixels finally being integrated in the resulting canonical spectral space to assign labels to each pixel.  
    
\section{Preliminary}\label{S3} 
For ease of exposition, mathematical background  is briefly outlined on t-SNE embedding~\cite{Maaten2008}  in Section~\ref{S3.1}  and canonical correlation analysis~\cite{Hardle2007} in Section~\ref{S3.2}.
\subsection{t-distributed stochastic neighbour embedding (t-SNE)}\label{S3.1}
 Given a set of $N$ high-dimensional $\boldsymbol{X} =   \{ \boldsymbol{x}_i\}, i=1,...,N$. As a tool for visualizing high-dimensional data by giving each data point a location in a two or three-dimensional map, the t-SNE~\cite{Maaten2008} aims to learn a very low $m$-dimensional map $\boldsymbol{Y} =\{\boldsymbol{y}_i\in \Re^m\}, i=1,...,N$, with $m$ typically chosen as 2 or 3.

The locations of the points  $ \boldsymbol{y}_i$ in the map are determined by minimizing the (non-symmetric) Kullback–Leibler divergence of the distribution 
of the pairwise similarity probabilities $P$ (of  $\boldsymbol{X}$) from the distribution $Q$  (of  $\boldsymbol{Y}$):
\begin{equation}\label{eq:1}
KL(P\|Q)=\sum_{i\neq j} \log \frac{p_{ij}}{q_{ij}}
\end{equation}
where $p_{ij}$ is joint probabilities of similarity $\{\boldsymbol{x}_j,  \boldsymbol{x}_i\}$,  $q_{ij}$ is joint probabilities of similarities of $\{\boldsymbol{y}_j, \boldsymbol{y}_i\}$.

To this end, the similarity of two distinctive $\boldsymbol{x}_j$ and  $\boldsymbol{x}_i$
is the conditional probability, $p_{j|i}$, defined as:
\begin{equation}\label{eq:2}
 p_{j|i} =\frac{\exp\left(-\|\boldsymbol{x}_i-\boldsymbol{x}_j \|^2   /(2\sigma_i^2) \right)}{\sum_{k\neq i} \exp\left(-\|\boldsymbol{x}_i-\boldsymbol{x}_k \|^2 /(2\sigma_i^2) \right) }
\end{equation}
for $i\neq j$. Set $p_{i|i}=0$, and 
 \begin{equation}\label{eq:3}
 p_{ij}=\frac{ p_{j|i}+ p_{i|j}}{2N}
 \end{equation}
By doing this, the distances between high dimensional data points are transferred into probabilities using a Gaussian distribution.

In t-SNE,  a Student t-distribution with one degree of freedom (which is the same
as a Cauchy distribution) is the heavy-tailed distribution in the low-dimensional map. Set $q_{ii}=0$. The similarity of two distinctive $\boldsymbol{y}_j$ to  $\boldsymbol{y}_i$
is  defined as:
\begin{equation}\label{eq:4}
 q_{ij} =\frac{ \left(1+\|\boldsymbol{y}_i-\boldsymbol{y}_j \|^2  \right)^{-1}  }{\sum_k\sum_{l\neq k}  \left(1+ \|\boldsymbol{y}_k-\boldsymbol{y}_l \|^2   \right)^{-1}}
\end{equation}
for $i\neq j$. $\boldsymbol{y}_i$ in the map are determined using gradient descent algorithm, with a user-set hyperparameter using the concept of perplexity, which is then used to determine $\sigma_i$ in (\ref{eq:2}) (please refer   to ~\cite{Maaten2008} for more details).

\subsection{Canonical correlation analysis}\label{S3.2}
 Canonical correlation analysis (CCA) is a way of measuring the linear relationship
between two sets of multidimensional variables. Consider two sets of random variables  $\boldsymbol{\Phi}=   \{ \boldsymbol{\phi}_i\}$,   $i=1, ..., d_1$   and  $\boldsymbol{\Psi}=   \{ \boldsymbol{\psi}_i \},  i=1, ..., d_2$  with zero mean. It is assumed that $\boldsymbol{\Phi}$, $\boldsymbol{\Psi}$ are full rank, and $d= \min(rank(\boldsymbol{\Phi}),rank(\boldsymbol{\Psi}))$.

The total covariance matrix
\begin{equation}\label{eq:5}
C=\left [ \begin{array}{cc}
 C_ {\boldsymbol{\Phi} \boldsymbol{\Phi}} & C_ {\boldsymbol{\Phi} \boldsymbol{\Psi}}\\
    C_ {\boldsymbol{\Psi} \boldsymbol{\Phi}} &     C_ {\boldsymbol{\Psi} \boldsymbol{\Psi}}      \end{array}\right] = E\left [ \left (  \begin{array}{c}
   \boldsymbol{\Phi}   \\
   \boldsymbol{\Psi}  \end{array}\right) \left (  \begin{array}{c}
   \boldsymbol{\Phi}   \\
   \boldsymbol{\Psi}  \end{array}\right)^{\rm T} \right]
\end{equation}

Define a set of two projection matrices $ \boldsymbol{A}=[\boldsymbol{a}_1,..., \boldsymbol{a}_d] \in \Re^{d_1\times d}$ and  $ \boldsymbol{B}=[\boldsymbol{b}_1,..., \boldsymbol{b}_d] \in \Re^{d_2\times d}$, which generate a set of linear combinations named $\boldsymbol{U}= \boldsymbol{\Phi}\boldsymbol{A}= [ \boldsymbol{u}_1, ...\boldsymbol{u}_d ]$   
and $\boldsymbol{V}= \boldsymbol{\Psi}\boldsymbol{B}= [ \boldsymbol{v}_1, ...\boldsymbol{v}_d]$. Each member of  $\boldsymbol{U}$ is paired with a member of  $\boldsymbol{V}$, as a set of canonical variables pairs $(\boldsymbol{u}_i, \boldsymbol{v}_i)$. 

The task in CCA is to find $ \boldsymbol{A},\boldsymbol{B}$ such that the correlations $\rho_i(\boldsymbol{u}_i, \boldsymbol{v}_i)$ are maximized. Represent
 \begin{equation}\label{eq:6}
 \rho_i= \frac{ \boldsymbol{a}_i^T C_ {\boldsymbol{\Phi} \boldsymbol{\Psi}} \boldsymbol{b}_i }{\sqrt{\boldsymbol{a}_i^T C_ {\boldsymbol{\Phi} \boldsymbol{\Phi}} \boldsymbol{a}_i}\sqrt{\boldsymbol{b}_i^T C_ {\boldsymbol{\Psi} \boldsymbol{\Psi}}\boldsymbol{b}_i}}, \ \ \ i=1,...,d.
 \end{equation}
Equivalently, 
   \begin{equation}\label{eq:7}
\max_{\boldsymbol{a}_i, \boldsymbol{b}_i} \rho_i=   \boldsymbol{a}_i^T C_ {\boldsymbol{\Phi} \boldsymbol{\Psi}} \boldsymbol{b}_i,  \forall  i
 \end{equation}
 subject to $\boldsymbol{a}_i^T C_ {\boldsymbol{\Phi} \boldsymbol{\Phi}} \boldsymbol{a}_i=1$, $ \boldsymbol{b}_i^T C_ {\boldsymbol{\Psi} \boldsymbol{\Psi}} \boldsymbol{b}_i =1$. 
 
To obtain the CCA solution, initially define~\cite{Hardle2007}
 \begin{equation}\label{eq:8}
{\cal K}=C_ {\boldsymbol{\Phi} \boldsymbol{\Phi}}^{-1/2}C_ {\boldsymbol{\Phi} \boldsymbol{\Psi}}C_ {\boldsymbol{\Psi} \boldsymbol{\Psi}}^{-1/2}
 \end{equation}
and perform singular value decomposition of  ${\cal K}$  as
\begin{equation}\label{eq:9}
{\cal K}= \Gamma \Lambda \Delta^{T}
\end{equation}
 with $\Gamma=[\gamma_1,...,\gamma_d]$,  $\Delta=[\delta_1,...,\delta_d  ]$.  $\Lambda=\text{diag}\{\lambda_1^{1/2},...,\lambda_d^{1/2}\}$.
 and $\lambda_1\ge \lambda_2 \ge ...\lambda_d$ are the nonzero eigenvalues of ${\cal K}^T{\cal K}$. $\gamma_i$ and $\delta_i$
are the left and right eigenvectors of  ${\cal K}$. 

Now define 
\begin{align}\label{eq:10}
&\boldsymbol{a}_i=C_ {\boldsymbol{\Phi}\boldsymbol{\Phi}} ^{-1/2} \gamma_i \nonumber\\
&\boldsymbol{b}_i=C_ {\boldsymbol{\Psi}\boldsymbol{\Psi}} ^{-1/2} \delta_i 
\end{align}
so that
\begin{equation}\label{eq:11}
cov( \boldsymbol{u}_i, \boldsymbol{u}_j) =  \boldsymbol{a}_i^{T}   C_ {\boldsymbol{\Phi} \boldsymbol{\Phi}}\boldsymbol{a}_j = \gamma_i^{T} \gamma_j =\left\{ \begin{array}{cc}
   1 &   i=j \\
  0&  i \neq j
\end{array} \right.
\end{equation}
\begin{equation}\label{eq:12}
cov( \boldsymbol{v}_i, \boldsymbol{v}_j) =  \boldsymbol{b}_i^{T}   C_ {\boldsymbol{\Psi} \boldsymbol{\Psi}}\boldsymbol{b}_j = \delta_i^{T} \delta_j =\left\{ \begin{array}{cc}
   1 &   i=j \\
  0&  i \neq j
\end{array} \right.
\end{equation}
and the correlation between $\boldsymbol{u}_i$ and $\boldsymbol{v}_i$ has maximal of
\begin{equation}\label{eq:13}
\rho( \boldsymbol{u}_i, \boldsymbol{v}_i) =   \gamma _i^T \Gamma \Lambda \Delta^{T} \delta_i  = \lambda_i^{1/2}.
\end{equation}

\section{Method}\label{S4}
This section starts with detailing the proposed pixel based  multiple spatial features generated from multi-band land cover images in Section \ref{S4.1}.  The proposed semi-supervised RBF-CCA algorithm for segmentation is presented in Section \ref{S4.2} based on t-SNE embedding produced from the full image feature space, as well as  a few  semantic labels.
 \subsection{Pixel based  multiple spatial features generation}\label{S4.1}
 Given a  multi-band  remote sensing image $I^{full} \in \Re^{H \times W \times M}$, where $H, W, M$ denote the height, width and number of grayscaled bands in $I^{full}$.  Without loss of generality, each grayscaled image is  denoted as $I_m$, $m=1,...,M$ and each pixel is denoted as $I_m(x,y)$.
\subsubsection{Cell feature vector}
A cell $C_{x,y}$  is simply a square patch centered at the current pixel, $I(x,y)$. For example, a 3 by 3 cell is  
\begin{equation}\label{eq:14}
C_{x,y} =\left[ \begin{array}{ccc}
 I(x-1,y-1)   & I(x-1,y)  & I(x-1,y+1)\\
I(x,y-1)   & I(x,y)  & I(x,y+1) \\
 I(x+1,y-1)   & I(x+1,y)  & I(x+1,y+1)\\
\end{array}\right]
\end{equation}
For a predetermined cell size $N_c$ by $N_c$, a cell feature vector $\boldsymbol{f}^{cell}(x,y) \in \Re^{N_c^2} $ is simply to reshape the patch centered at $I (x,y)$ into a vector. Note that if a pixel is near the edge of $I_m$, the associated cell matrix $C_{x,y}$ is  created by an enlarged image from $I_m$ with auxiliary pixels. This is done by mirroring the image along the edge to the outside, so that the image is symmetrical with across its edge. The cell size $N_c=7$ is used in this work.

\subsubsection{Local binary pattern feature vector}\label{S4.2}
Similarly,  a square patch $Pat(x,y)$ centered at the current pixel $I (x,y)$  is considered for generating local binary pattern (LBP) features \cite{Ojala1994}.  For each pixel in the patch, compare the pixel to each of its 8 neighbours $I (x_i,y_i)$ as indexed by $i=1,...,8$, on its top, left-top, left-middle, left-bottom, bottom, right bottom, right middle, right-top, which follows the pixels along a complete anti-clockwise circle. The resulting LBP value is
\begin{equation}\label{eq:15}
LBP(x,y) =\sum_{i=1}^8 Id\left[ I(x_i,y_i)\ge I(x,y) \right] * 2^{i-1}
\end{equation}
where the indicator function $Id(S)$  is defined as one if the statement $S$ is true, zero otherwise. 
$LBP(x,y)$ is an integer ranged with [$0,  2^8-1$].
 %
 Furthermore, motivated by the fact that only a subset of 58 binary patterns, named uniform LBP patterns, accounts for the majority of LBP patterns, the set of uniformed LBP patterns is reduced from 256 to 58 + 1 = 59, where  all non-uniformed LBP patterns  merges as one additional pattern~\cite{Ojala2001}.
 
 Over a patch, the histogram of the uniformed  LBP  can be generated, which may readily  be used  as features for pixel $I(x,y)$. In order to obtain more LBP samples for histogram generation, we use a larger patch size $ N_t > N_c$ to obtain texture features. In this work, a $11$  by $11$  patch ($N_t=11$) is used to create $\boldsymbol{f}^{LBP}(x,y) \in \Re^{59}$.   

\subsubsection{Gray-Level Co-occurrence Matrix statistical feature vector (GLCM)}
A co-occurrence matrix~\cite{Haralick1973} is a matrix that is defined over an image as the distribution of co-occurring pixel values at a given offset. In order to generate pixel based feature of GLCM, the same square patch $Pat(x,y)$  centered at the current pixel $I (x,y)$  as used in LBP feature generation is adopted.  
Given the $N_t$ by $N_t$ gray-leveled   $Pat(x,y)$ with $L$  different pixel levels (e.g. $L=8$),  the co-occurrence matrix computes how often pairs of pixels with a specific value and offset occur in the image. Specifically, the  co-occurrence matrix $G =\{g(i,j)\}$ is calculated in relation to a given offset, where
\begin{align}\label{eq:16}
&g(i,j)= \nonumber \\
& \sum_{x=1}^{N_t} \sum_{y=1}^{N_t}  Id\left[  Pat(x,y)= i  \ \ \texttt{and} \ \  Pat(x+\delta x,y+\delta y)= j \right]
\end{align}
 where $i, j \in [1,...,L]$ are pixel gray levels. $[x, y]$ is the spatial position of a pixel and  [$\delta x, \delta y$] is the offset, defining the spatial relation for which this matrix is calculated.

 The  statistics properties \cite{Bevk2002} feature vectors, including contrast, correlation, energy and homogeneity, $ \boldsymbol{f}^{GLCM}=[ f_1^{GLCM}, f_2^{GLCM},f_3^{GLCM},f_4^{GLCM}]^T$  can be derived from the  GLCM as
 \begin{align}\label{eq:17}
&f^{GLCM}_1=\frac{1}{L^2} \sum_i\sum_j |i-j|^2 g(i,j)\\
&f^{GLCM}_2=\frac{1}{L^2}  \sum_i\sum_j (i-\mu_r)(j-\mu_c)\frac{g(i,j)}{\sigma_r\sigma_c} \\
&f^{GLCM}_3=\frac{1}{L^2} \sum_i\sum_j   g(i,j)^2\\
&f^{GLCM}_4= \frac{1}{L^2} \sum_i\sum_j \frac{ g(i,j)}{1+|i-j|}
\end{align}
 where $\mu_r,\mu_c$ are the row and column mean of the marginal probabilities of GLCM given by
 \begin{align}\label{eq:18}
& \mu_r=\frac{1}{L^2}  \sum_i i \sum_j g(i,j)\\
&\mu_c= \frac{1}{L^2} \sum_j  j \sum_i g(i,j)  
\end{align}
 and $\sigma_r,\sigma_c$ are associated standard deviation. 
 \begin{align}
& \sigma_r= \frac{1}{L} \sqrt{\sum_i (i -\mu_r)^2 \sum_j g(i,j)}\\
&\sigma_c= \frac{1}{L} \sqrt{\sum_j (j -\mu_c)^2 \sum_i g(i,j)}
\end{align}  
 
\subsubsection{Multiple spatial features}
 The process of spatial feature generation maps each pixel in a gray level image $I_m$ to 
\begin{equation}\label{eq:19}
 I_m(x,y) \in \Re \rightarrow  \boldsymbol{f}(x,y) \in \Re^{N_{c}^2+ N_{LBP}+N_{GLCM}}
\end{equation}
where $$\boldsymbol{f}(x,y) =[ [ \boldsymbol{f} ^{cell}(x,y)]^T,  [ \boldsymbol{f} ^{LBP}(x,y)]^T,  [\boldsymbol{f} ^{GLCM}(x,y) ]^T ]^T.$$ 

Over all pixels of $I_m$,    we have 3D tensor
\begin{equation}\label{eq:20}
\boldsymbol{X}_m =\{\boldsymbol{f}(x,y) \}\in \Re^{H \times W \times (N_{c}^2+ N_{LBP}+N_{GLCM})}
\end{equation}
with $N_{LBP}=59$,  $N_{GLCM}=4$,  $\boldsymbol{X}_m$  is then reshaped as a feature matrix as
\begin{equation}\label{eq:21}
 \boldsymbol{X}_m =[ \boldsymbol{f}_1, ...,  \boldsymbol{f}_{N} ]^T  \in \Re^{N \times (N_c^2+ N_{LBP}+N_{GLCM}) } 
\end{equation}
where $i=1, ...N$, $N=H   W$ is the total number of pixels per image in any band $m \in [1, ..., M] $.    

For a multi-band image $I^{full} \in \Re^{H \times W \times M}$, this process is then repeated for all bands, as illustrated in  Figure \ref{Figure:1}. For remote sensing applications, an additionally derived band may be created based on physical properties. For example, an $NDVI$ (Normalized Difference Vegetation Index) is computed as the difference between near-infrared (IR) and red (RED) reflectance divided (element wise) by their sum.
\begin{equation}\label{eq:22}
NDVI= \frac{IR-R}{IR+R}
\end{equation}
where $IR$, $R$ denote near-infrared, and red bands. In this study, we used the NDVI as an additional band to generate of multiple features from NDVI imagery. LIDAR data are also employed in the feature space where available, with its first echo (FE), last echo (LE), and intensity, in which LIDAR FE and LE correspond to the first and last points from where the laser beam is reflected, and LIDAR intensity represents the laser pulse amplitude of the LIDAR FE. When the surface height information of DSM (Digital Surface Model) is given, it is also used as a band in the feature space. 
The final feature matrix is 
\begin{equation}\label{eq:23}
\boldsymbol{X}=[\boldsymbol{X}_1,  \boldsymbol{X}_2, ....,  \boldsymbol{X}_M] \in \Re^{N \times n}
\end{equation}
 $n = M \times (N_c^2+ N_{LBP}+N_{GLCM}) $ is a total number of features for the full multi-band remote sensing image $I^{full}$.

\subsection{Proposed semi-supervised spectral clustering algorithm using RBF-CCA}\label{S4.2}
Suppose that  we  have some $N_l\ll N_u$ labelled data points, as $D_{train}=\{  \boldsymbol{x}_i,t_i \}_{i=1}^{N_l}$, where $ \boldsymbol{x}_i= [x_{i,1} ,...,  x_{i,n}]^{\rm T} \in \Re^n$ is a high dimensional input feature vector. $t_i \in \{1,...K\}$ for a given $K \ll N_l$. Let the input data points with labels be denoted as $\boldsymbol{X}^{(L)}=[ \boldsymbol{x} _1,..., \boldsymbol{x}_{N_l}]^{\rm T}$. Simultaneously, there are $N_u \gg N_l$ unlabelled data points given as $\boldsymbol{X}^{(U)}=[\boldsymbol{x}_{N_l+1} ,..., \boldsymbol{x}_{N}]^{\rm T}$.  Denote the completed input data points
$$\boldsymbol{X}=\left[ \begin{array}{c}
     \boldsymbol{X}^{(L)} \\  \boldsymbol{X}^{(U)}
\end{array} \right]  \in \Re^{N \times n},$$ 
$N=N_u+N_l$. The goal of semi-supervised clustering  is to partition these points into $K$ disjoint sets, with partial assistance of supervised learning from the data set $D_{train}$. 

Initially, we apply t-SNE algorithm  for dimensionality reduction $\boldsymbol{x}_i \in \Re^n \longrightarrow  \boldsymbol{y}_i \in \Re^m$. Denote 
$\boldsymbol{Y}=\text{t-SNE} (\boldsymbol X)$ as the full input features.

\begin{figure}[!ht]
   \centering
    \includegraphics[width=0.5\textwidth]{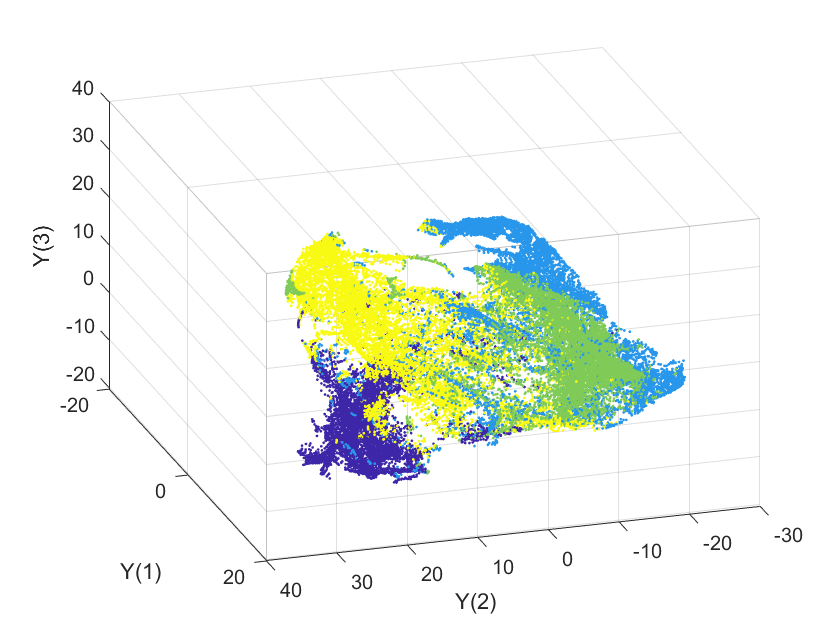}  
     \caption{Exemplary 3D view of t-SNE embedding of top, right sub-image (first data set TopoSys GmbH RS, see Section \ref{S5}), which shows irregular shaped clusters.}
    \label{Figure:2}
\end{figure}

It is known that the t-SNE algorithm can generate  clusters for visualisation, however,
if a conventional clustering algorithm, e.g. K-means, is applied to assign cluster labels to $\boldsymbol{y}_i$, it may not be very effective, due to that the clustering criterion in k-means is over simplified, by just minimizing average distances to each cluster centre, and is not suitable for irregularly shaped clusters, as illustrated in Figure \ref{Figure:2}. An improvement would be to have a universal approximation model that is capable of modelling arbitrary cluster shapes, with which to make much appropriate clustering decision. This is the motivation for this work. 

The radial basis functions are a cornerstone in approximation theory~\cite{PoggioT1990}. The output of a radial basis function reduces as the distance between the input and its fixed centre  increases, giving its ability to locally identify new data to these centres. In order to model the irregularly shaped t-SNE clusters,   we propose to use a modified CCA based on radial basis function using data set $D_{train}$, referred to as RBF-CCA.  Specifically, all the training data  $D_{train}$, which are assigned ground truth labels, are set to be RBF centres.  For convenience, denote the embedding of labelled  data points as $\boldsymbol{c}_j=\boldsymbol{y}_j$, $i=1, ..., N_l$. Over $\boldsymbol Y$, let   $\phi_{i,j}=\exp\left(-\frac{\|\boldsymbol{y}_i -\boldsymbol{c}_j \|^2}{2\sigma^2}  \right)$, in which  $\boldsymbol{c}_j$ are the centres of radial basis functions, $\sigma$ is a preset proper hyperparameter. In this work, we simply use 
\begin{equation}
\sigma= \sqrt{  \frac{1}{NN_l} \sum_{i=1}^{N} \sum_{j=1}^{N_l} \|\boldsymbol{y}_i -\boldsymbol{c}_j \|^2 }
\end{equation}
to construct a matrix 
\begin{equation}\label{eq:24}
\boldsymbol{\Phi}^{full} =\left[ \begin{array}{ccc}
\phi_{1,1}& \cdots&\phi_{1,N_l}\\
 \vdots& \ddots&\vdots\\
\phi_{N,1}& \cdots&\phi_{N,N_l}\\
 \end{array} \right]
\end{equation}
followed by mean removal as $$\boldsymbol{\Phi}^{full} \leftarrow \boldsymbol{\Phi}^{full} -  \text{mean}(\boldsymbol{\Phi}^{full}).$$  Let 
\begin{equation}\label{eq:25}
 \boldsymbol{\Phi}^{full}  =\left[ \begin{array}{c}
     \boldsymbol{\Phi}  \\  \boldsymbol{\Phi}^{U} 
\end{array} \right],
\end{equation}
where  $ \boldsymbol{\Phi} \in \Re^{N \times n_l} $ is based on labelled data $D_{train}$, which is used as the first set of variables in our proposed RBF-CCA. By examining  (\ref{eq:10}), we can see that this fills up all pair-wise RBF between a query (input features in t-SNE) to that of training data (with known labels). At each row, only data points close to any given labelled training data will be excited, otherwise their values are close to zero.  Since it is expected that visible clusters are seen in t-SNE space,  this means that most of the data query may excite some cluster via its local RBF centres. We then relate the second set of variables in CCA from ground truth in $D_{train}$.   The second variable CCA is designed as follows.

In order to create the second set of variables with the objective of semi-supervised clustering,  $\boldsymbol{\Psi}\in \Re^{N_l \times K}$ is generated as 
 \begin{equation}\label{eq:26}
\boldsymbol{\Psi} =\left[ \begin{array}{ccc}
\psi_{1,1}& \cdots&\psi_{1,K}\\
 \vdots& \ddots&\vdots\\
\psi_{N_l,1}& \cdots&\psi_{N_l,K}\\
 \end{array} \right]  
\end{equation}
in which
\begin{equation}\label{eq:27}
\psi_{i,j}=\left\{ \begin{array}{cc}
   1 &   j =t_i \\
  0&  j \neq  t_i 
\end{array} \right.
\end{equation}
followed by mean removal as $$\boldsymbol{\Psi}  \leftarrow \boldsymbol{\Psi}  -  \text{mean}(\boldsymbol{\Psi}). $$
Clearly using canonical correlation analysis (CCA)  for the two sets of multidimensional variables   $\boldsymbol{\Phi}$, $\boldsymbol{\Psi}$, as defined above, aims to capture the relationship between the nonlinear RBF transform based on t-SNE clusters and the cluster labels for the labelled data set $D_{train}$. The proposed algorithm, as shown in Algorithm \ref{ALG1},    is based on Section \ref{S3.1} to apply the two sets of random vectors as defined above, referred to as RBF-CCA.  Note that Line 5 returns the   canonical variable  corresponding to the full input features, which are used for spectral clustering. The proposed  semi-supervised spectral clustering based on RBF-CCA  is given in Algorithm \ref{ALG2}.
\begin{algorithm}[ht]
\caption{Modified CCA using radial basis functions (RBF-CCA).}  
\label{ALG1}
\begin{algorithmic}[1]
\REQUIRE Number $K$  of clusters to construct; Labelled data $D_{train}=\{  \boldsymbol{x}_i,t_i \}_{i=1}^{N_l}$;  Unlabelled data points $\boldsymbol{X}^{(U)}=[\boldsymbol{x}_{N_l+1} ,..., \boldsymbol{x}_{N}]^{\rm T} $. Complete data in space of \text{t-SNE} $\boldsymbol{Y}$ (see Section \ref{S3.1}). 
\STATE Construct   $\boldsymbol{\Phi}^{full}$ and  $\boldsymbol{\Psi}$ using (\ref{eq:24}) and (\ref{eq:26}) respectively, then remove their mean.
\STATE Recover  $\boldsymbol{\Phi}$ using (\ref{eq:25}).
\STATE Perform CCA to obtain $ \boldsymbol{A} \in \Re^{N_l \times K}$ and $ \boldsymbol{B}\in \Re^{K \times K}$ (see Section \ref{S3.2})
\STATE  Calculate $\boldsymbol{U}^{full}= \boldsymbol{\Phi}^{full}\boldsymbol{A}$ using  complete input data set.
\STATE  Return $\boldsymbol{U}^{full} =\{ u_{i,j} \} \in \Re^{N \times K}$.
\end{algorithmic}
\end{algorithm}

\begin{algorithm}[ht]
\caption{Proposed semi-supervised algorithm based on  RBF-CCA and t-SNE} 
\label{ALG2}
\begin{algorithmic}[1]
\REQUIRE Number $K$  of clusters to construct; Labelled data $D_{train}=\{  \boldsymbol{x}_i,t_i \}_{i=1}^{N_l}$;  Unlabelled data points $\boldsymbol{X}^{(U)}=[\boldsymbol{x}_{N_l+1} ,..., \boldsymbol{x}_{N}]^{\rm T} $.   
\STATE Form complete input data matrix $\boldsymbol{X}$.  Apply $\boldsymbol{Y}=\text{t-SNE} (\boldsymbol X)$ to each point.
\STATE  Call Algorithm 1.
\STATE  Form the matrix $\boldsymbol{Z}=\{z_{i,j}\} \in \Re^{N \times K}$ by normalising the rows to norm one, i.e. to set
\begin{equation}\label{eq:28}
z_{i,j}=u_{i,j}\bigg/\sqrt{\sum_{j=1}^{K}  u_{i,j} ^2}.
\end{equation}
\FOR{$i=1,...,N$}
\STATE Let $\hat{\boldsymbol{z}}_i  \in \Re^{K}$ be the vector corresponding to the $i$th  row of $\boldsymbol{Z}$.
\ENDFOR
\STATE Cluster the points $\hat{\boldsymbol{z}}_i$, $i=1,...,N$ with the $k$-means algorithm~\cite{Haykin2009} into clusters $C_1,...,C_K$.
 \STATE  Return:  Find clusters $k \in \{1,...,K\}$ with $\{k,  \hat{\boldsymbol{z}}_i  \in  C_k\}$ and assign original data points $\boldsymbol{x}_i$  according to cluster's index set of  $k=1,...,K$.
\end{algorithmic}
\end{algorithm} 

Since canonical variables, which are related to eigenvectors, are applied for clustering,   this approach is similar to spectral clustering algorithms~\cite{UlrikeLuxburg2007}. Note that the idea of Lines 7 and 8 in Algorithm \ref{ALG2} is borrowed from a spectral clustering algorithm~\cite{Ng2001} in which clusters in the space of normalized eigenvectors of input data are formed. Here, the clusters in the space of covariates of input data are to be formed. The idea is similar to spectral clustering in that both employ orthogonal spectral analysis of input features, and perform clustering using projected variables in a new orthogonal space. Note that according to (\ref{eq:11}), these  canonical variables are orthogonal over the small labelled data set, to maximize cross correlation with the labelled output. Different to spectral clustering,  the proposed algorithm, as a semi-supervised approach,  realizes transfer learning  via projection of the associated canonical variables to unlabelled data sets. It is therefore reasonable to assume that if the labelled data set is randomly sampled and representative of the full data set, this could lead to good clustering performance over the full data set.
 
 The computational complexities of each part of the proposed algorithm is in the order of $O(nN^2)$  (t-SNE), which is further scaled by iterations of gradient descent algorithms,  $O(N_l^3)$ (RBF-CCA), and $O(N)$ (k-means clustering). Since $N\gg N_l$, the main cost is due to obtaining the t-SNE embedding, which is a drawback for problems with large $N$ and $n$.  To mitigate this, principal component analysis (PCA) can be applied for reducing $n$, and the simple strategy of divide and conquer can be employed to reduce $N$. In our experiments, images are divided into sub-images, with final segmentation combined at the end.
  
\begin{table*}[ht]
\caption{Example 1: Comparison of segmentation accuracy (percentage of correctly identified pixel labels). Three variants of CCA are based on linear, polynomial and radial basis function of t-SNE features.}\label{Table1}  
\centering
\begin{tabular}{|l|l|c|c|c| c|c|c|   } \hline 
 \multicolumn{2}{|c| }{ Data sets }   &   \ \ \ \ \ k-means on \ \ \ \ \ \   &  k-means on   &Linear & Polynomial     & Proposed    \\  
 \multicolumn{2}{|l| }{   }  & multiple features  &     t-SNE embedding   & CCA  &CCA& RBF-CCA\\ \hline  
 &  Sub-image (top, left)  & $62.04$    
 &  $71.22$ &$74.79\pm 0.26$  &$77.90\pm 0.46$  & $87.28\pm 0.25$  \\  
            TopoSys GmbH       &   Sub-image (top, right) &   $51.34$  &  $  72.97$ &$72.98\pm0.16 $  & $73.99\pm0.17 $ &   $84.50\pm0.21$   \\  
                  (First data set)                 &   Sub-image (bottom, left) &  $58.80$   &  $74.01$ & $76.01\pm0.19 $  & $79.99\pm0.21 $   &  $87.88\pm0.42 $  \\ 
                               &   Sub-image (bottom, right) &   $55.54$ &  $71.70$ &$76.77\pm0.10$ &$79.37\pm0.68$ & $87.16\pm0.19$  \\ \hline  \hline 
TopoSys GmbH & \multirow{2}*{Full image} &  \multirow{2}*{$60.52$}   &  \multirow{2}* {$72.91$}  &\multirow{2}* {$71.04 \pm 0.53$}    &\multirow{2}* {$75.64 \pm 0.17$}  &  \multirow{2}* {$84.77\pm0.20$ }\\ (Second data set) &&&&&&\\\hline 
 \end{tabular}
 \end{table*}

\section{Experiments}\label{S5}
Two sets of experiments with multi-band remotely sensed images,  which include hand-labelled ground truth,  are conducted for validating the proposed algorithm. For  training purposes, we repeatedly used five percent of ground truth as $N_l$ labeled data, which are randomly selected.  The ground truth are also used for the evaluation purpose.  In order to obtain the final performance metrics, the predicted cluster labels are mapped into the given ground truth via the well known Kuhn-Munkres algorithm~\cite{lovász2009}.
 
 \begin{table}[ht]
\centering
\caption{Intersection over union (IOU) of Example 1 (TopoSys GmbH). }\label{Table2}  
  \begin{tabular}{|l|c| c|  } \hline   
  \multirow{2}*{Class}      &   First data set  & Second data set  \\  
    &  mean IOU &   mean IOU \\\hline
 1-  Building    & $0.9536$   &  $0.9371$   \\
 2-  Tree & $0.9050$ & $0.9096$\\
 3- Low   vegetation & $0.8723$  &$0.8688$ \\
 4- Impervious surface OR Car & $0.9054$ &$0.8769$ \\ \hline
  \end{tabular} 
 \end{table} 

{\bf Example 1}: Two data sets obtained from Trimble Holdings GmbH (formerly TopoSys GmbH, Germany) are experimented. Each data set contains seven bands, which are LIDAR first echo FE, last echo LE, colour image red (R), green (G), and blue (B),  near infrared (IR), and LIDAR intensity (IN). An additional band of NDVI was generated, so eight bands are used for each data set.
 
\begin{figure*}[!ht]
   \centering
      (a) \hspace{0.2\textwidth} (b) \hspace{0.2\textwidth} (c) \hspace{0.2\textwidth}  (d)\\
    \includegraphics[width=0.2\textwidth]{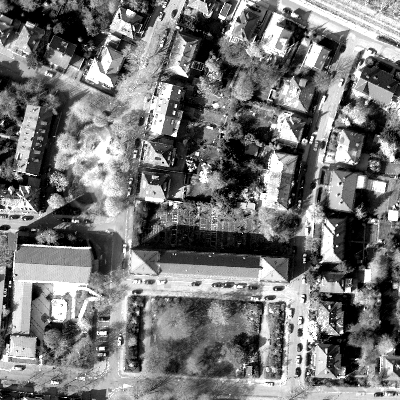} \ \  \ \includegraphics[width=0.2\textwidth]{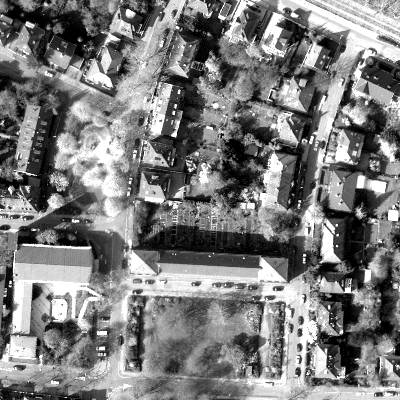} \ \  \  \includegraphics[width=0.2\textwidth]{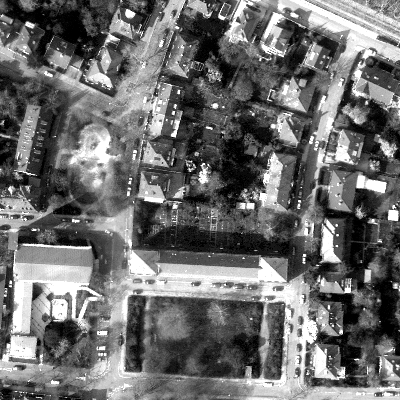} \ \ \ \includegraphics[width=0.2\textwidth]{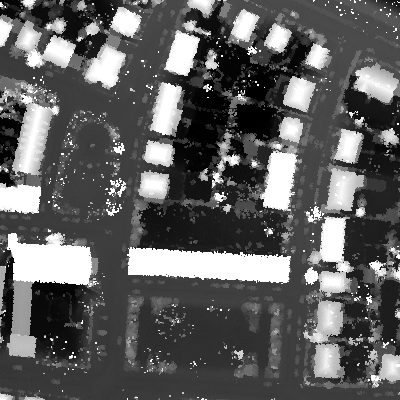} \\
 (e) \hspace{0.2\textwidth} (f) \hspace{0.2\textwidth} (g)\hspace{0.2\textwidth} (h) \\
    \includegraphics[width=0.2\textwidth]{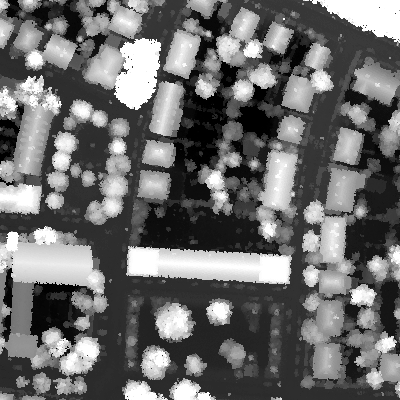} \ \  \ \includegraphics[width=0.2\textwidth]{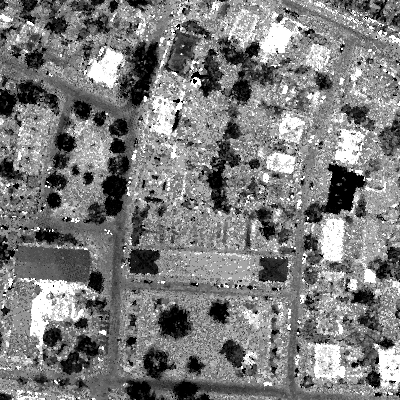} \ \ \ 
    \includegraphics[width=0.2\textwidth]{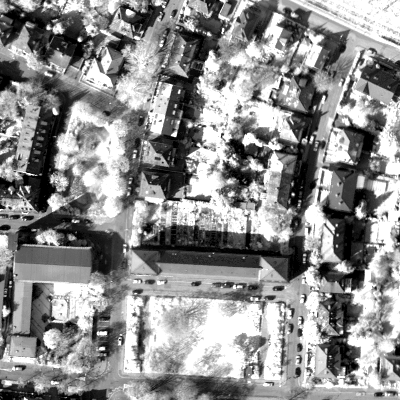}  \ \ \    \includegraphics[width=0.2\textwidth]{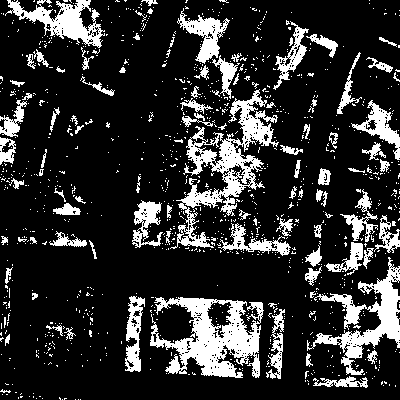} \\
    \caption{Eight multispectral images   of TopoSys GmbH (the first data set, Example 1) (a) Red; (b) Green; (c) Blue;  (d) Last echo; (e) First echo; (f) LIDAR intensity;   (g) Near infrared;  (h)   NDVI. }
    \label{Figure:5}
\end{figure*}

\begin{figure*}[!ht]
   \centering
      (a) \hspace{0.2\textwidth} (b) \hspace{0.2\textwidth} (c) \hspace{0.2\textwidth} (d)\\
  \includegraphics[width=0.2\textwidth]{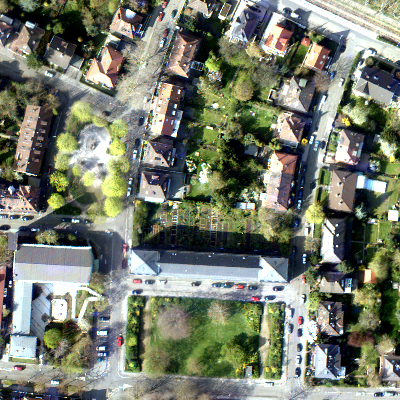} \ \   \includegraphics[width=0.2\textwidth]{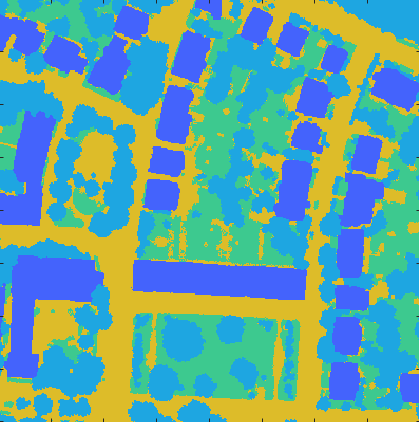} \ \   \includegraphics[width=0.2\textwidth]{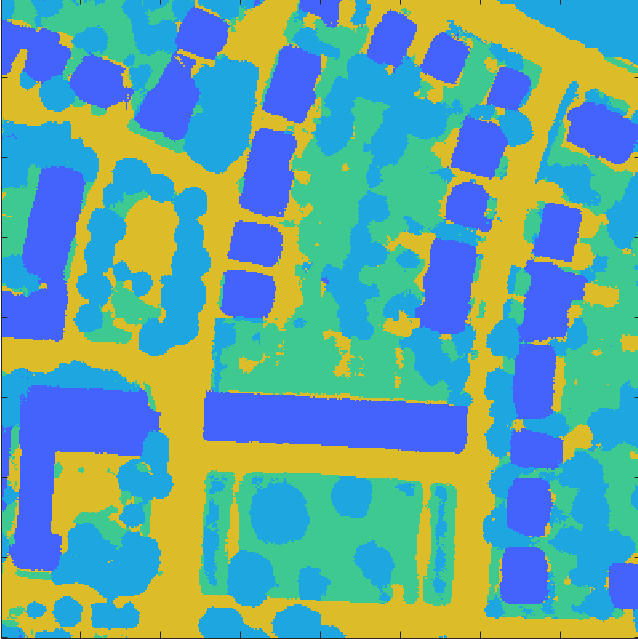} \ \    \includegraphics[width=0.22\textwidth]{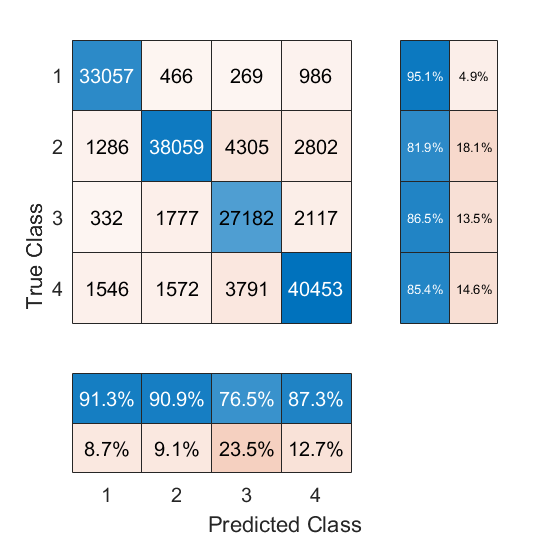}  \\
     \caption{Segmentation results of  TopoSys GmbH (the firts data set, Example 1): (a) RGB view (b) Ground truth; (c) Segmentation; and  (d) Confusion matrix (1- Building;  2 -Tree ; 3 - Low   vegetation ; and   4-  Impervious surface OR Car.) }
    \label{Figure:6}
\end{figure*}

 \textit{$\bullet$  The first data set}   has a size of $400 \times 400$ pixels per band, with seven measured data plus the derived NDVI, as shown in Figure \ref{Figure:5}, and the RGB view of the image is shown in Figure \ref{Figure:6}(a). The data set was then  processed by splitting into four  $200 \times 200$ sub-images, each of which is treated as  $I^{full}\in \Re^{200\times 200 \times 8}$. This results in   $N=40000$, $n=896$, for each  sub-image (left top, right top, left bottom and right bottom). The segmentation was performed for each part respectively, and final results are obtained by merging the parts into the complete image. The ground truth of this image contains four classes (1- Building;  2 -Tree ; 3 - Low   vegetation ; and   4-  Impervious surface OR Car) that were manually labelled pixel-wise, of which  only 5\% was  used for training, and all are used for validation. In analysis of the data set, there is higher presence of \lq \lq Low   vegetation'' that needs to be segmented, and the ground truth has a type \lq\lq Impervious surface OR Car'', which makes   data class population well-balanced.  The segmentation accuracies of an average of ten random experiments of the proposed RBF-CCA algorithm are included, in comparison with a few baseline approaches, as shown in Table \ref{Table1}.   The k-means algorithm was applied in original high dimensional data space, as well as in 3D t-SNE space.  For the two additional versions of CCA used for the purpose of comparative studies, they differ from RBF-CCA only in how to form  $\boldsymbol{\Phi}^{full}$. In the linear CCA algorithm,  the t-SNE features are used without any nonlinear transformation. For the polynomial  CCA algorithm,  the t-SNE features together with their quadratic terms are applied.   It can be seen that although k-means clustering based on t-SNE embedding offers improvement over clustering of original data space, the proposed RBF-CCA has much improved clustering performance over both unsupervised k-means algorithm for original data   and  t-SNE embedding data. The proposed RBF-CCA algorithm is better than linear CCA and polynomial CCA, both are included to demonstrate the superior approximation capability of RBF to model the highly nonlinear t-SNE clusters.  The resulting segmented four sets of 200 by 200  pixels labels are then combined. The ground truth,   segmentation representation and classification confusion matrix are shown in Figure \ref{Figure:6}(b)-(d) respectively, also  demonstrating excellent performance  for the first data set.

\begin{figure*}[!ht]
   \centering
        (a) \hspace{0.2\textwidth} (b) \hspace{0.2\textwidth} (c) \hspace{0.2\textwidth}  (d)\\
    \includegraphics[width=0.2\textwidth]{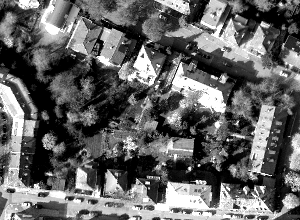} \ \  \ \includegraphics[width=0.2\textwidth]{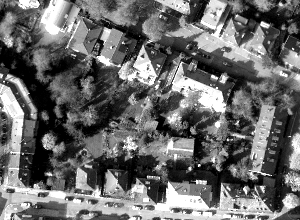} \ \  \  \includegraphics[width=0.2\textwidth]{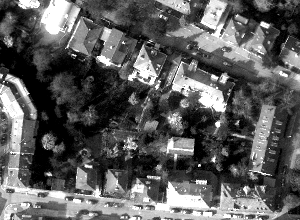} \ \ \ \includegraphics[width=0.2\textwidth]{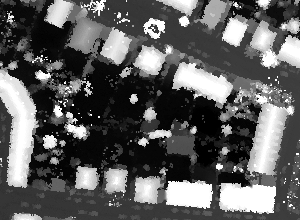} \\
 (e) \hspace{0.2\textwidth} (f) \hspace{0.2\textwidth} (g)\hspace{0.2\textwidth} (h) \\
    \includegraphics[width=0.2\textwidth]{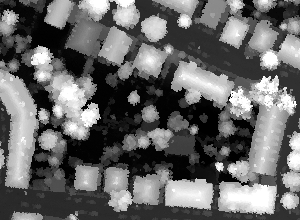} \ \  \ \includegraphics[width=0.2\textwidth]{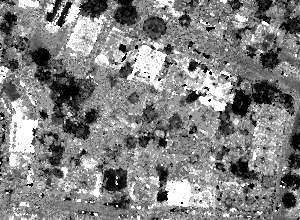} \ \ \ 
    \includegraphics[width=0.2\textwidth]{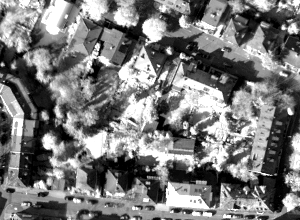}  \ \ \    \includegraphics[width=0.2\textwidth]{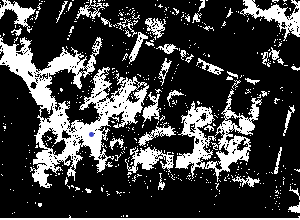} 
    \caption{Eight multispectral images   of TopoSys GmbH  (the second data set, Example 1) (a) Red; (b) Green; (c) Blue;  (d) Last echo; (e) First echo; (f) LIDAR intensity;   (g) Near infrared;   and (h) NDVI}
    \label{Figure:7}
\end{figure*}

\begin{figure*}[!ht]
   \centering
      (a) \hspace{0.2\textwidth} (b) \hspace{0.2\textwidth} (c) \hspace{0.2\textwidth} (d)\\
  \includegraphics[width=0.21\textwidth] {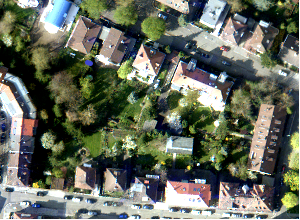} \ \   \includegraphics[width=0.22\textwidth]{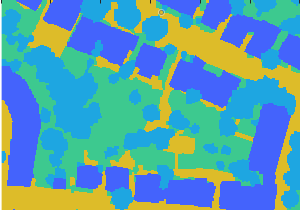} \ \   \includegraphics[width=0.22\textwidth]{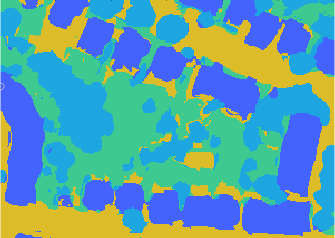} \ \    \includegraphics[width=0.18\textwidth]{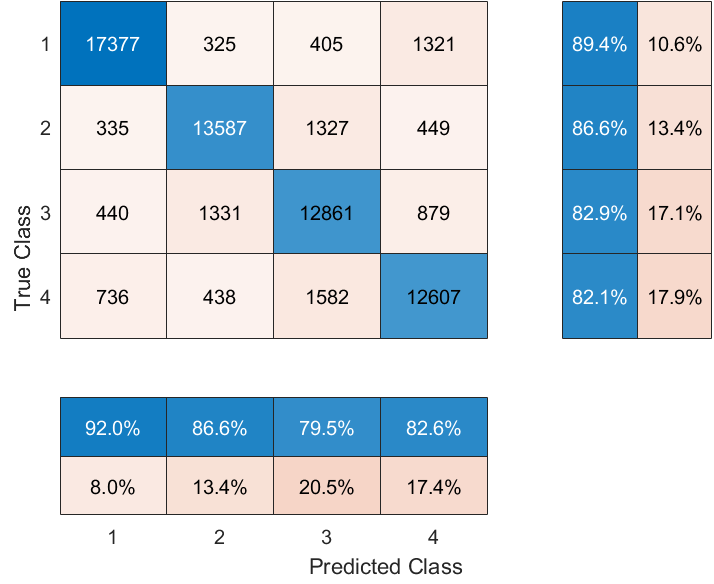}  \\
     \caption{Segmentation results of the TopoSys GmbH (the second data set, Example 1); (a) RGB view (b) Ground truth; (c) Segmentation; and  (d) Confusion matrix ((1- Building;  2 - Tree ; 3 - Low   vegetation ; and   4-  Impervious surface OR Car. ) }
    \label{Figure:8}
\end{figure*}

\textit{$\bullet$  The second data set}  is measured in $220 \times 300$ per band, as shown in Figure \ref{Figure:7}.  In this case, the data set was  processed  as a whole, so $I^{full}\in \Re^{220 \times 300 \times 8}$, resulting in  $N=66000$, $n=896$.  A coloured RGB  is shown in Figure \ref{Figure:8}(a).  Similar to the first data set, four classes for segmentation are labelled as (1- Building;  2 -Tree ; 3 - Low   vegetation ; and   4-  Impervious surface OR Car).   The ground truth,  segmentation representation and classification confusion matrix are shown in Figure \ref{Figure:8}(b)-(d) respectively.  The  segmentation accuracies, as also included in Table \ref{Table1}, demonstrate superior performance for all classes in comparison with the k-means algorithm, over multiple features and t-SNE embedding, as well as linear CCA  and polynomial CCA.

The quality of segmentation of Example 1 over both data sets, as measured by intersection over union (IOU) from set theory, is shown in Table \ref{Table2}, where the mean IOU is calculated by averaging IOU of one against  all other clusters involved.


\begin{figure*}[!ht]
   \centering
      (a) \hspace{0.2\textwidth} (b) \hspace{0.2\textwidth} (c) \\
    \includegraphics[width=0.2\textwidth]{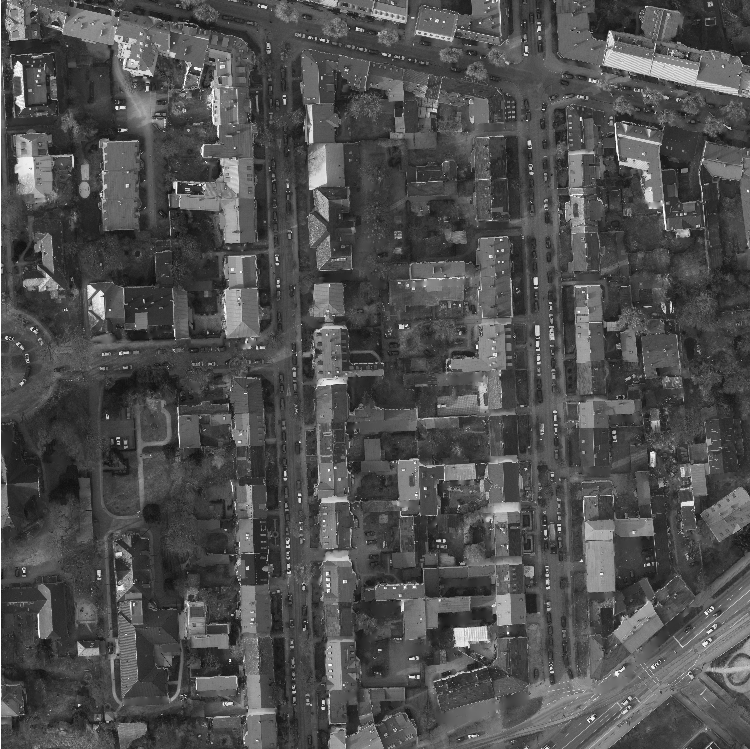} \ \  \ \includegraphics[width=0.2\textwidth]{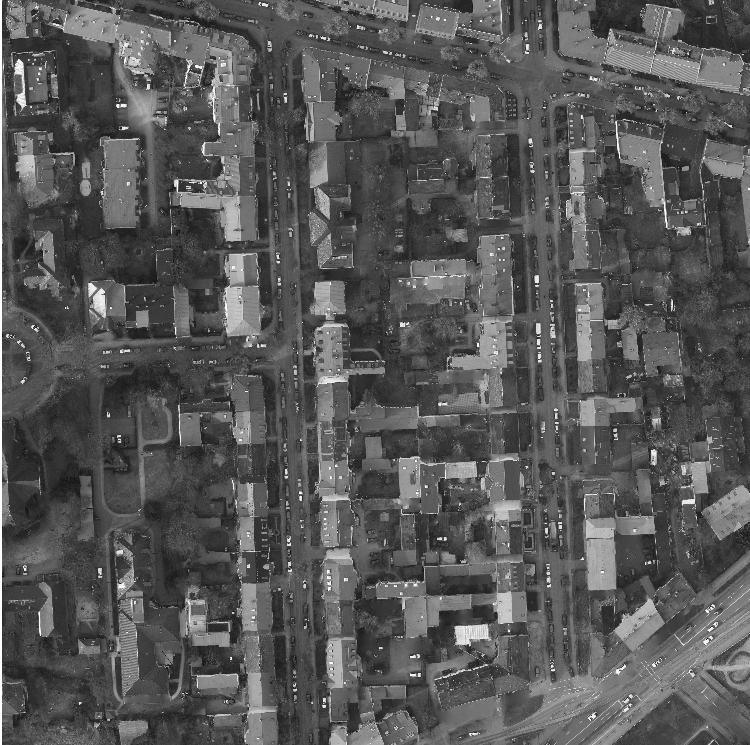} \ \  \  \includegraphics[width=0.2\textwidth]{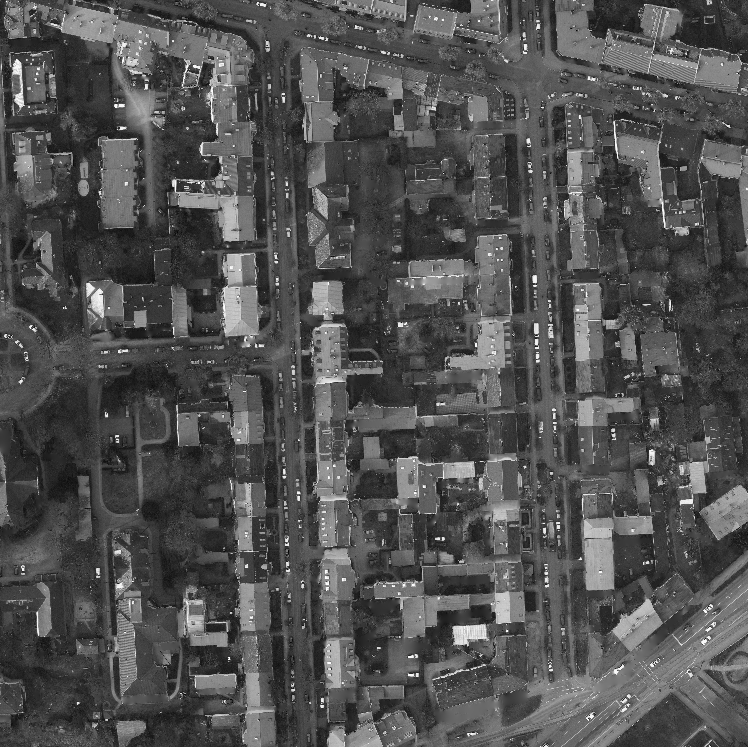}  \\
  (d)\hspace{0.2\textwidth}  (e)    \hspace{0.2\textwidth}  (f)   \\
 \includegraphics[width=0.2\textwidth]{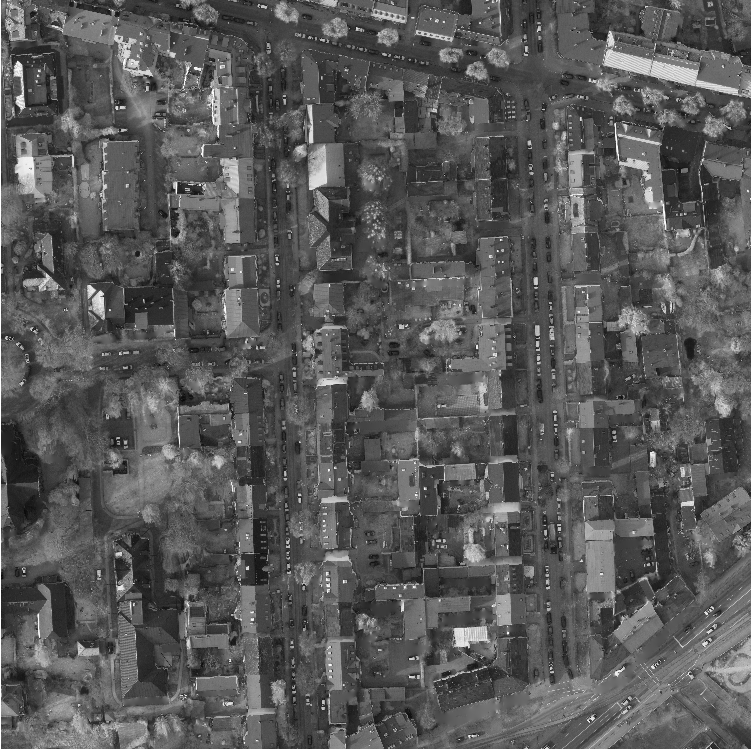} \ \ \  \includegraphics[width=0.2\textwidth]{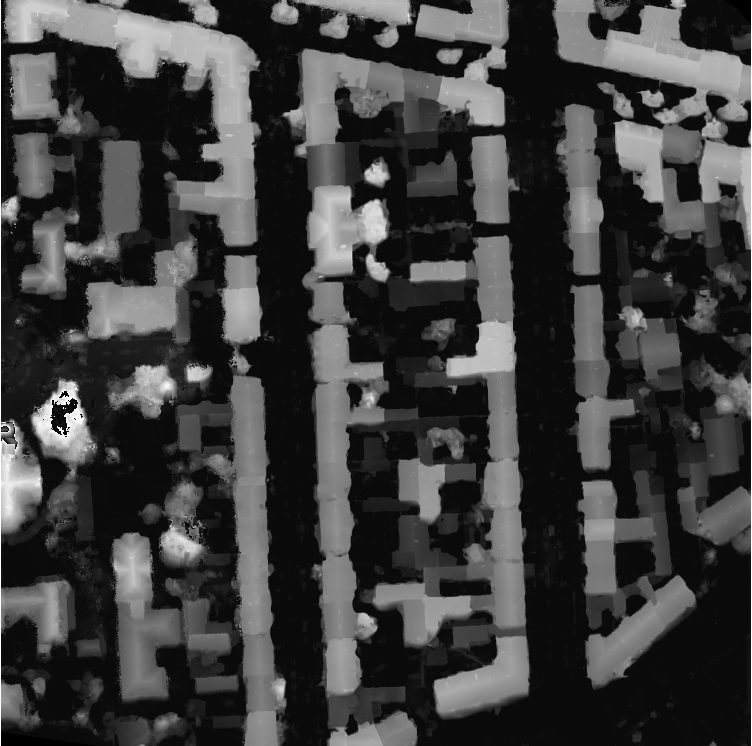}  \ \  \ \includegraphics[width=0.2\textwidth]{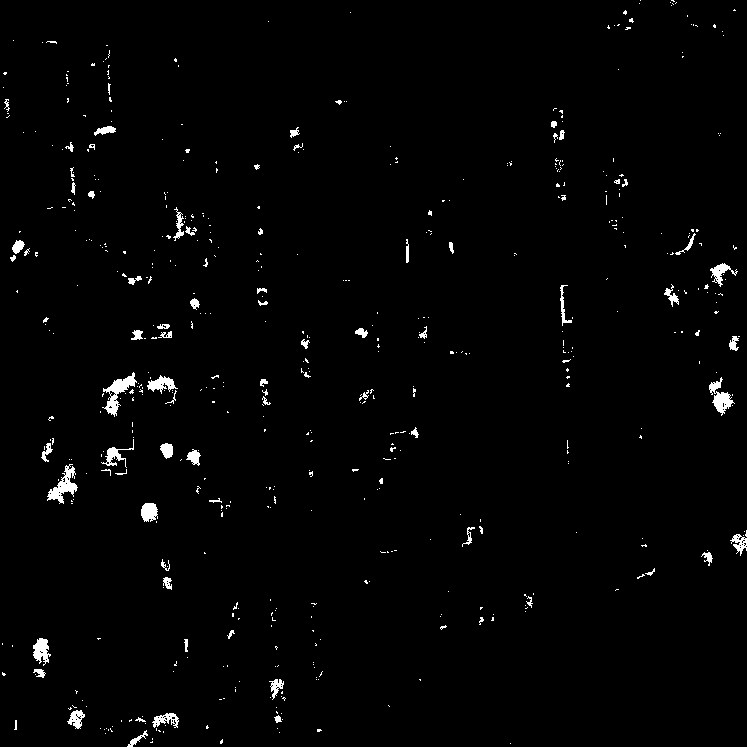} \\
    \caption{ Six multispectral images channels of Potsdam7-8 (Example 2) (a) Red; (b) Green; (c) Blue;  (d) Infra-red; (e) DSM;  (f) NDVI  }
    \label{Figure:9}
\end{figure*}
 \begin{figure*}[!ht]
   \centering
     (a) \hspace{0.2\textwidth} (b) \hspace{0.2\textwidth} (c) \hspace{0.2\textwidth} (d)\\
  \includegraphics[width=0.2\textwidth,height=0.2\textwidth]{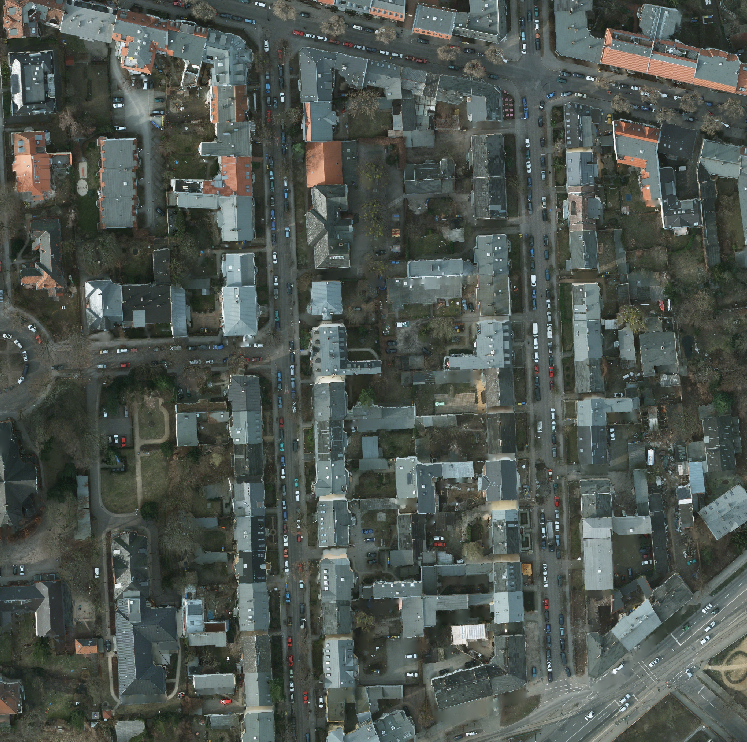} \ \   \includegraphics[width=0.2\textwidth,height=0.2\textwidth]{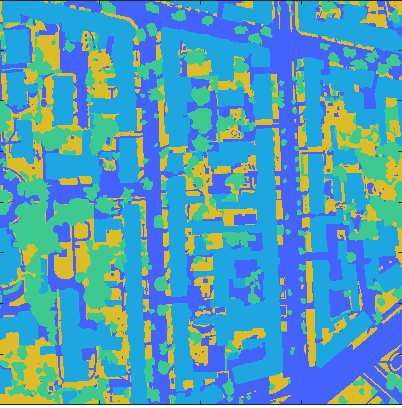} \ \   \includegraphics[width=0.2\textwidth,height=0.2\textwidth]{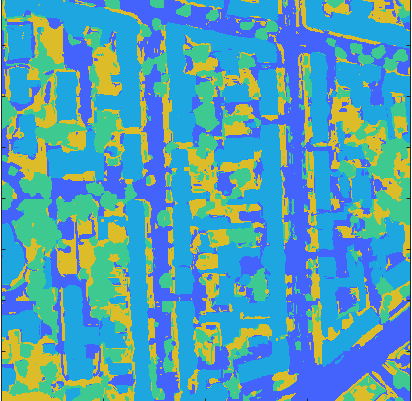} \ \    \includegraphics[width=0.22\textwidth]{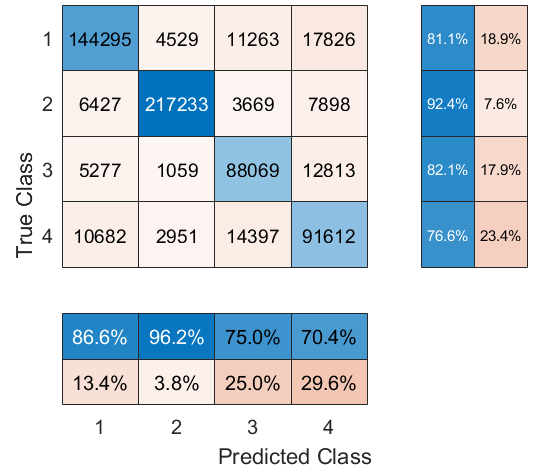}  \\
     \caption{ Segmentation results  of Potsdam7-8 data set (Example 2); (a) RBG view;  (b) Ground truth; (c) Segmentation; and  (d) Confusion matrix (1- Building;  2 - Impervious surface OR  Car OR Clutter; 3 - Tree ; 4- Low  vegetation) }
    \label{Figure:10}
\end{figure*}

{\bf Example 2}: A data set showing City of Potsdam in ISPRS 2D Semantic Labelling Contest is used in this experiment \cite{Potsdam}. The website  contains 38 patches of the same size. The patch 7-8 was used to demonstrate the effectiveness of the proposed approaches. The original data set has a high resolution 6000 by 6000 pixels, with multi-band images consisting of R, G, B, IR and normalized DSM. Together with an additional band of NDVI, six bands are as shown in Figure \ref{Figure:9} and the coloured RGB view is shown in Figure \ref{Figure:10}(a).   The high resolution multi band images are resized to 800 by 800 pixels,  which are then divided into sixteen 200 by 200 pixels sub-images,  so that  $I^{full}\in \Re^{200\times 200 \times 8}$, based on which the proposed algorithm RBF-CCA is performed.

For  each of 16 sub-images,  $N=40000$ and  $n = 672$. The original ground truth was Impervious surfaces (RGB: 255, 255, 255), Building (RGB: 0, 0, 255), Low vegetation (RGB: 0, 255, 255), Tree (RGB: 0, 255, 0), Car (RGB: 255, 255, 0)
 Clutter/background (RGB: 255, 0, 0). This is also resized to 800 by 800 pixels, followed by rounding to the closest (255 or 0) in each R, G, B band. The ground truth of this image are set as four classes (1- Building;  2 - Impervious surface OR  Car OR Clutter; 3 - Tree ; 4- Low  vegetation), as shown in Figure \ref{Figure:10}(b).  The segmentation accuracies of all sixteen sub-images, each of which the average results of ten random experiments (using $N_l=5\% N$ labelled data in training of RBF-CCA) are listed as shown in Table \ref{Table3}, by comparing with the k-means algorithm on multiple features and t-SNE embedding and the two variants of CCAs. It can be seen linear CCA fails due to inability to model nonlinearity in t-SNE clusters, and the polynomial CCA has only limited improvement over linear CCA since it cannot model nonlinearity about labelled data samples locally.  Both linear and polynomial CCA have similar performance to unsupervised clustering approaches.  Owing to the universal ability of RBFs to approximate t-SNE embeddings, the proposed RBF-CCA algorithm  has the best segmentation performance. 
 
\begin{table*}[ht]
\caption{Example 2: Comparison of segmentation accuracy (percentage of correctly identified pixel labels). Three variants of CCA are based on linear, polynomial and radial basis function of t-SNE features.}\label{Table3}  
\centering
\begin{tabular}{|l|l|c|c|c| c|c|c|   } \hline   \multicolumn{2}{|c| }{ Data sets }   &   \ \ \ \ \ k-means on \ \ \ \ \ \   &  k-means on   & Linear  & Polynomial & Proposed    \\  
 \multicolumn{2}{|l| }{   }  & multiple features  &     t-SNE embedding   & CCA& CCA & RBF-CCA \\ \hline  
   \multirow{4}*{Potsdam patch 7-8}  & Sub-image (top, left)     & $55.94$ &  $53.39$& $50.12\pm0.01$  &$64.21\pm0.67$ & $85.63\pm0.52$  \\ 
    \multirow{4}*{(top, left quarter)} & Sub-image (top, right)  & $67.94$   &  $62.03$  & $57.42\pm0.44$   &$73.35\pm0.42$ &    $88.31\pm 0.20$ \\ 
       &Sub-image (bottom, left)  &  $59.15$   &  $75.81$   &  $71.61\pm 1.98$  & $75.63\pm 0.25$  &   $84.55\pm 0.33$  \\ 
     & Sub-image (bottom, right)&   $66.68$ & $65.52$  &  $65.05\pm 0.13$& $68.33\pm 0.31$&   $83.31\pm 0.43$ \\\hline  \hline
        \multirow{4}*{Potsdam patch 7-8}  &Sub-image (top, left)  &  $63.10$   & $63.96$&$60.51\pm0.16$ &$64.78\pm0.79$&   $83.68\pm0.60$\\ 
    \multirow{4}*{(top, right quarter)} & Sub-image (top, right)  &   $68.80$ & $64.37$    &$53.85\pm0.86$  &$64.06\pm0.41$ &   $87.16\pm0.52$  \\ 
     &  Sub-image (bottom, left)  &  $60.17$  &  $55.64$   &$62.33\pm0.36 $& $68.95\pm 0.38$&   $82.54\pm0.37$ \\ 
     & Sub-image (bottom, right)&  $70.94$  & $63.38$   &$55.50\pm0.88$ &$66.94\pm1.28$&   $81.38\pm0.57$  \\ \hline \hline
        \multirow{4}*{Potsdam patch 7-8}  & Sub-image (top, left) &  $65.14$  &  $62.63$  &$60.73\pm0.34$ &$74.63\pm0.35$&  $87.07\pm0.33$  \\ 
   \multirow{4}*{(bottom, left quarter)}  &Sub-image (top, right)   &  $53.64$   &  $51.92$   &$50.63\pm 0.01$ &$65.29\pm0.50$& $84.59\pm0.52$  \\ 
       &   Sub-image (bottom, left) &   $68.23$ &  $47.57$ &$48.01\pm0.14$ &$72.27\pm0.27$&  $87.00\pm0.17$  \\ 
     &Sub-image (bottom, right) & $57.24$  &  $52.32$ &$49.59\pm0.88$ &$59.24\pm2.05$&  $83.93\pm0.40$  \\\hline  \hline
        \multirow{4}*{Potsdam patch 7-8}  & Sub-image (top, left) &  $58.86$   &  $58.52$  &$53.46\pm0.12$ &$61.12\pm4.36$& $81.98\pm0.59$   \\ 
   \multirow{4}*{(bottom, right quarter)}  & Sub-image (top, right)   &  $72.37$   &   $56.29$ &$52.72 \pm0.27$ &$55.20\pm0.39$&  $82.11\pm0.66$  \\ 
       &   Sub-image (bottom, left) &  $65.28$  &  $62.25$  &$54.52\pm0.91$ &$59.58\pm0.72$&  $83.14\pm0.76$ \\ 
     &Sub-image (bottom, right) &  $73.64$  &   $62.04$ &$59.46\pm0.43$ &$57.56\pm2.66$&  $86.35\pm 0.45$ \\ \hline 
 \end{tabular}
 \end{table*} 
 
The resulting segmented sixteen 200 by 200 pixels labels are combined and is shown in Figure \ref{Figure:10}(c), and the pixel based classification confusion matrix is shown in Figure \ref{Figure:10}(d). Finally, the mean IOU is shown  in Table \ref{Table4}, demonstrating that the proposed approaches are highly effective.  
  \begin{table}[ht]
  \centering
\caption{Intersection over union (IOU) of Example 2 (ISPRS 2D Potsdam Patch 7-8 ). }\label{Table4}  
  \begin{tabular}{|l|c|    } \hline 
 \multirow{2}*{Class}   & \multirow{2}*{mean IOU}\\& \\ \hline 
 1-  Building    &    $0.8872$ \\
 2-  Impervious surface    & \multirow{2}*{$0.9610$}  \\   \ \ \ OR  Car OR Clutter &   \\
 3- Tree &   $0.8516$  \\
 4- Low  vegetation &  $0.8093$  \\ \hline
  \end{tabular} 
 \end{table} 

\section{Conclusions}\label{S6}
 Applied to  image segmentation of remotely sensed urban areas, this paper has introduced a novel semi-supervised  spectral clustering  method based on multiple texture features of multispectral images plus LIDAR data where available.   Centered at each raw pixel in multiple bands, including the inferred vegetation index band, a pixel patch is vectorized, into  concatenated image texture features of LBP and GLCM to be represented as a data point in a high-dimensional feature space, which are mapped to a 3-D visually t-SNE embedding space.  Using a small subset of pixels having segmentation labels,  this paper has extended canonical correlation analysis with RBF function to nonlinear CCA.  We have proposed to perform t-SNE embedding initially, based on which RBF-CCA algorithm is then developed.  The first set  of random variables is designed based on a set of RBFs using the labelled   data points in t-SNE embedding, and the second set of random variables are related to corresponding labels. The proposed algorithm learns the associated projection matrix from RBF-CCA, followed by learning the associated canonical variables for the  full multi-band image. Finally, the canonical variables  are  clustered using k-means clustering.  It is shown that the proposed algorithm is capable of excellent segmentation for several land cover data sets, despite only using a small amount of labelled data. 

In future work with regard to semi-supervised multispectral image segmentation for remote sensing imagery, different land cover scenarios will be considered. The algorithm will be further developed  and optimized in terms of robustness in challenging situations with a focus on imbalanced clusters, scattered important objects, and segmentation of similar objects. It will also be extended to deal with sea surfaces covered by ice, water mixed with ice, and shadowed by clouds from satellite multispectral images.

 \bibliographystyle{IEEEtran}


\end{document}